\documentclass[10pt,twocolumn,letterpaper]{article}

\usepackage{iccv}
\usepackage{times}
\usepackage{epsfig}
\usepackage{graphicx}
\usepackage{amsmath}
\usepackage{amssymb}
\usepackage{multirow}

\usepackage{times}
\usepackage{epsfig}
\usepackage{graphicx}
\usepackage{amsmath}
\usepackage{amssymb}
\usepackage{multirow}
\usepackage{booktabs}
\usepackage[T1]{fontenc}
\usepackage{CJKutf8}
\usepackage[english]{babel}
\usepackage{subfig}

\usepackage{array}
\makeatletter
\newcommand{\thickhline}{%
    \noalign {\ifnum 0=`}\fi \hrule height 1pt
    \futurelet \reserved@a \@xhline 
}
\newcolumntype{!}{@{\hskip\tabcolsep\vrule width 1pt\tabcolsep\hskip}} 
\makeatother

\usepackage[breaklinks=true,bookmarks=false]{hyperref}

\iccvfinalcopy 

\def\iccvPaperID{****} 

\begin{document}

\title{
Propagated Perturbation of Adversarial Attack for well-known CNNs: Empirical Study and its Explanation
}

\author{Jihyeun Yoon \thanks{Equal contribution, alphabetical order}\\
LG CNS\\
{\tt\small jihyeun.yoon@lgcns.com}
\and
Kyungyul Kim \footnotemark[1]\\
LG CNS\\
{\tt\small kyungyul.kim@lgcns.com}
\and
Jongseong Jang \thanks{Corresponding author} \\
LG Science Park \\
{\tt\small j.jang@lgsp.co.kr}
}

\maketitle

\begin{abstract}
Deep Neural Network based classifiers are known to be vulnerable to perturbations of inputs constructed by an adversarial attack to force misclassification. Most studies have focused on how to make vulnerable noise by gradient based attack methods or to defense model from adversarial attack. The use of the denoiser model is one of a well-known solution to reduce the adversarial noise although classification performance had not significantly improved. In this study, we aim to analyze the propagation of adversarial attack as an explainable AI(XAI) point of view. Specifically, we examine the trend of adversarial perturbations through the CNN architectures. To analyze the propagated perturbation, we measured normalized Euclidean Distance and cosine distance in each CNN layer between the feature map of the perturbed image passed through denoiser and the non-perturbed original image. We used five well-known CNN based classifiers and three gradient-based adversarial attacks. From the experimental results, we observed that in most cases, Euclidean Distance explosively increases in the final fully connected layer while cosine distance fluctuated and disappeared at the last layer. This means that the use of denoiser can decrease the amount of noise. However, it failed to defense accuracy degradation.
\end{abstract}


\section{Introduction}
In the computer vision field, deep neural networks(DNNs) achieve successful performance across various areas such as image classification, object detection, and semantic segmentation. 
But even though DNN is well trained, it can be easily degraded when noise is added to input data. 
Especially, DNN models trained by gradient-descent and back-propagation can be deteriorated by gradient-based noise attack, so called adversarial noise \cite{barreno2010security,nguyen2015deep,sharif2016accessorize}.
In such an adversarial attack to classifier case, noise is located near discriminant hyperplane of DNN models, which makes easy to deceive the classifier. Thus, it is accomplished by making noise in a vulnerable area of DNN \cite{szegedy2013intriguing, kurakin2016adversarial}.
Briefly, adversarial noise is a practical method because it could perturb target DNN without involvement in the learning process, and it often happens that it is difficult to visually confirm the presence of noise.

As a defense method for the type of adversarial attack, it is very natural to consider gradient masking, which hides the gradient of DNNs. But, gradient-based noise could be easily generated by substituting a model to a target classifier called a black-box attack \cite{papernot2016transferability, papernot2017practical}. 
Therefore, most differentiable DNN could be easily exposed to gradient-based attack.

One of basic defense method against adversarial noise is to remove adversarial perturbation before the classifier.
Among many denoising methods, denoising Auto-Encoder(DAE) \cite{vincent2008extracting} can be designed as a convolutional neural network (CNN) that can reduce the number of parameters and improve calculation efficiency while it maintains the performance \cite{gondara2016medical}.
For example, encoder-decoder structure and lateral skip-connection for residual learning based methods such as U-Net \cite{ronneberger2015u}, FusionNet \cite{quan2016fusionnet} and stacked U-Net \cite{sevastopolsky2018stack} were introduced. 
Though the schemes are designed for image segmentation, it could be trained as a denoiser using pixel distance based objective function.

As shown the table 1, the experiment addressed that the performance of  classifiers was not significantly improved in spite of reduced noise by the denoiser(FusionNet). 
It means that the characteristics of adversarial noise affect classification performance. 
For a good understanding of this phenomenon, adversarial noise has to be observed, how it would be propagated while it passes through DNN. 

\begin{table*}[t!]
\begin{center}
\begin{tabular}{l|ll|ll} 
\toprule
\multirow{2}{*}{adversarial mode} & \multicolumn{2}{l|}{Perturbation of Val. set(pix)}        & \multicolumn{2}{l}{Top-1 Val. Acc. (\%)}  \\ 
\cline{2-5}
                                  & MSE($x_{ori}^{val}$, $x_{adv}^{val}$) & MSE($x_{ori}^{val}$, DN($x_{adv}^{val}$)) & F($x_{adv}^{val}$) & F(DN($x_{adv}^{val}$))                 \\ 
\hline
FGSM V1                           & 0.166               & 0.268                   & 80.24     & 80.31                         \\
FGSM V2                           & 6.018               & 2.703                   & 27.06     & 45.47                         \\
i-FGSM V1                         & 1.071               & 0.689                   & 59.37     & 66.77                         \\
i-FGSM V2                         & 3.621               & 1.415                   & 29.6      & 45.68                         \\
mi-FGSM                           & 1.262               & 0.574                   & 76.43     & 78.41                         \\
\bottomrule
\end{tabular}
\end{center}
\caption{Adversarial perturbation measured by mean square error (MSE) and corresponding accuracy on a validation set of TinyImageNet. Classifier and denoiser are Inception-Resnet V2(train accuracy 86.67\%, validation accuracy 82.0\%) and FusionNet respectively. $x_{ori}^{val}$ is the original validation images and $x_{adv}^{val}$ is the corresponding adversarial examples. DN($\cdot$) is the output passing through the denoiser and F($\cdot$) is the classifier output. It seems that there is no exact dependency between the amount of noise and validation accuracy.}
\end{table*}

In this study, we examined of propagation behavior from input to output using well-known classifiers, three gradient-based attack noise \-- fast gradient sign method(FGSM) \cite{goodfellow2014explaining}, iterative fast gradient sign method(i-FGSM) \cite{kurakin2016adversarial} and momentum iterative fast gradient sign method(mi-FGSM) \cite{dong2018boosting}. 
We analyzed why this kind of noise is difficult to defence using denoiser and DNN classifier.


\subsection{Contribution}
In this study, we made the following contributions.
\begin{itemize}
\item We observed propagation of black-box and white-box adversarial attack noise from input to output layer for DNN models, and tried to explain how to generate hyperplane.
    We Also showed that efficacy of standard denoiser based on DAE is limited for adversarial defense. 
    According to the observation, we found that although perturbation is reduced, propagation is amplified to a similar level. 
    We also analyzed the difference between original and adversarial samples and provided understanding about defense against adversarial attack.
    \item We experimented using various CNNs which have different capacity. From the experiment, we provided insight about propagation behavior respect to capacity and architecture. 

\end{itemize}

\begin{figure*}[ht!]
\begin{center}
   \includegraphics[width=0.6\linewidth]{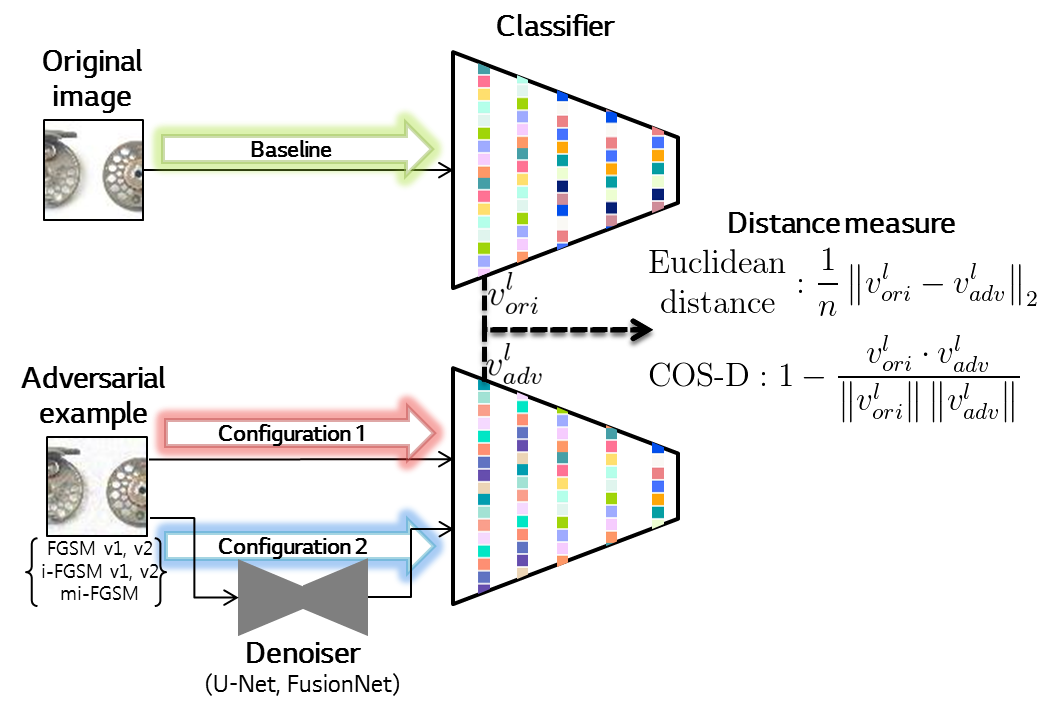}
\end{center}
   \caption{Concept of the experiment. We measure differences between feature maps of different inputs at the specified layer in the identical classifier. In the baseline setting, we feed original images to a classifier. Then, we also feed five types of adversaries and denoised adversaries to classifier for comparing the difference. Measuring difference is conducted by normalized Euclidean and cosine distance (NE-D \& COS-D) between two feature maps, $v^{l}_{ori}$ and $v^{l}_{adv}$ which are the feature maps for the original and adversarial inputs at the $l$ th layer. $n$ is the number of elements in the feature map.}
\label{fig:one}
\end{figure*}

\section{Preparations}
For observing the propagation of adversarial perturbation through a classifier, we calculated some distance between feature maps of original data and perturbed one layer by layer. Before running into it, three essential components, 1)generation of adversarial examples, 2)trained classifiers by training the dataset, and 3)trained denoisers by perturbed datasets, are required. 

\subsection{Generation of adversarial examples}
On generation adversarial examples, we used three types of gradient-based attack and generated five adversarial datasets. datasets are generated by using TinyImageNet \cite{noauthor_undated-jl} and black\&white-box attack \cite{papernot2016transferability, papernot2017practical} in this paper.
\begin{itemize}
    \item \textbf{\textit{Fast gradient sign method(FGSM)}} \cite{goodfellow2014explaining}: To generate adversarial example, the attacker accumulates perturbation to the direction of input-output gradient to the original image, as follows:
    \begin{equation}
    x_{adv} \xleftarrow{} x + \epsilon \cdot sign(\nabla _{x}J ( \theta, x, y))
    \end{equation}
    In the equation $x$ is an original input image, $y$ is the ground truth label, $\epsilon$ is the step size of the perturbation, $sign()$ is the sign function, $J ( \theta, x, y)$ is the loss function when $\theta$ is the trained attacker's parameters, and $\nabla _{x}$ is the gradient function for $x$. The attacker continues to accumulate it until it successes to mis-classify for the classifier or gets to $\epsilon$'s step limitation.
    
    \item \textbf{\textit{Iterative fast gradient sign method(i-FGSM)}} \cite{kurakin2016adversarial}: Similar to FGSM, this attack iteratively computes the direction of the gradient and accumulates it to the image perturbed just before, as follows:
    \begin{equation}
    x_{k+1} \xleftarrow{} x_{k} + \epsilon \cdot sign(\nabla _{x}J ( \theta, x_{k}, y))
    \end{equation}
    Attack would be stopped when it can lead misclassification for the classifier or gets to $\epsilon$'s step limit.
    
    \item \textbf{\textit{Momentum iterative fast gradient sign method(mi-FGSM)}} \cite{dong2018boosting}: In this method, a gradient is updated from the previous version with the momentum term, then it is accumulated to the image perturbed just before, as follows:
    \begin{align}
        \begin{aligned}
        g_0 &= 0, ~~~x_0 = \text{original image}\\
        g_{k+1} &\xleftarrow{} \mu \cdot g_k + \frac{\nabla _x J(\theta, x_{k}, y)}{||\nabla _x J(\theta, x_{k}, y)||_1}\\
        x_{k+1} &\xleftarrow{} x_k + \epsilon \cdot sign(g_{k+1})
        \end{aligned}
    \end{align}
    In the equation, $x_k$ is k times perturbed noise and $\mu$ is balancing coefficient to adjust the change of the gradient by using a previous one. It also iterates this procedure until the classifier misclassifies the input or it gets to $\epsilon$'s step limit.
\end{itemize}

In general, the attacker is a substitute model of the classifier. In the black box attack, attack for input is iteratively conducted until a classifier returns a wrong label. During the attack, as an input-output gradient, adversarial perturbation is calculated in the attacker, which is another classifier trained on the same task. Based on these methods, we generated five adversarial examples sets by using Foolbox library \cite{DBLP:journals/corr/RauberBB17} which is a python toolbox of large collection about the adversarial attack. The configurations for generating adversarial examples are shown in Table 2. 

\begin{table}[ht!]
\begin{center}
\small
\begin{tabular}{l|l|l} 
\hline
          & \begin{tabular}[c]{@{}l@{}}attacker\\(top-1 val. acc.) \end{tabular} & \begin{tabular}[c]{@{}l@{}}classifier\\(top-1 val. acc.) \end{tabular}  \\ 
\hline
FGSM v1   & \multirow{5}{*}{\begin{tabular}[c]{@{}l@{}}ResNet-18 \\(60\%) \cite{DBLP:journals/corr/HeZRS15}\end{tabular}}                                     & ResNet-18 (60\%)                                                         \\
FGSM v2   &                                                                      & Inception-ResNet V2 (80\%)                                               \\
i-FGSM v1 &                                                                      & ResNet-18 (60\%)                                                         \\
i-FGSM v2 &                                                                      & Inception-ResNet V2 (80\%)                                               \\
mi-FGSM   &                                                                      & Inception-ResNet V2 (80\%)                                               \\
\hline
\end{tabular}
\caption{Configurations for generating adversarial datasets.}
\end{center}
\end{table}

\subsection{Preparation of classifiers trained by training set}
To observe the propagation of adversarial perturbation, 5 well-known CNNs with different capacity, i.e. VGG-19 \cite{DBLP:journals/corr/SimonyanZ14a}, ResNet V2-50 \cite{DBLP:journals/corr/HeZR016}, Inception-ResNet V2 \cite{DBLP:journals/corr/SzegedyIV16}, DenseNet-201 \cite{DBLP:journals/corr/HuangLW16a} and SENet-154 \cite{DBLP:journals/corr/abs-1709-01507}, were trained with TinyImageNet training set. For generalization, we trained them nearly perfectly on the training set. Before training, parameters of all classifiers were initialized by pre-trained model on ImageNet \cite{ILSVRC15} and input images were resized to 128$\times$128 (original size is 64$\times$64) by bilinear interpolation. As an aside, we slightly modified their architecture (by adjustment stride size in the low level layers), as pre-trained models were optimized to the image size (299$\times$299) of ImageNet. Training examples were sequentially transformed with random crop (range 0.85$\sim$1.0) and horizontal flip (prob. 0.5) for each epoch, and normalized by a range of (-1.0, 1.0). Each of them was trained with Adam optimizer \cite{DBLP:journals/corr/KingmaB14} with an initial learning rate of 1.0e$^{-5}$ and L2 regularization coefficient of 1.0e$^{-5}$. The prepared classifiers and their capacity are shown in Table 3. 

\begin{table}[h!]
\centering
\begin{tabular}{l|l|l|l} 
\toprule
                    & \begin{tabular}[c]{@{}l@{}}top-1\\training\\acc.(\%) \end{tabular} & \begin{tabular}[c]{@{}l@{}}Num.\\layers \end{tabular} & \begin{tabular}[c]{@{}l@{}}Num.\\parameters \end{tabular}  \\ 
\hline
ResNet V2-50        & 99.905                                                      & 50                                                    & 25.56M                                                     \\
VGG-19              & 99.221                                                      & 19                                                    & 138.36M                                                    \\
SENet-154           & 99.334                                                      & 154                                                   & 113.45M                                                    \\
Inception-ResNet V2 & 99.598                                                      & 234                                                   & 55.97M                                                     \\
DenseNet-201        & 99.907                                                     & 201                                                   & 18.47M                                                     \\
\bottomrule
\end{tabular}
\caption{well-known classifiers and their capacity. We used them to observe the propagation of adversarial perturbation. For generalization, all classifiers were nearly perfectly trained to the TinyImageNet training dataset.}
\end{table}

\subsection{Preparation of denoiser trained by adversarial example sets}
As a denoising architecture, the authors selected U-Net \cite{ronneberger2015u} and FusionNet \cite{quan2016fusionnet} as a denoiser. 
The reason is that U-Net methodology has proven to perform well
in maintaining the robustness of models against adversarial
attacks (in the NIPS2017 adversarial vision challenge \cite{nips-avc2018}).
So we have experimented with U-Net. Moreover, FusionNet
which is an improved version of U-Net by skip-connections
was used \cite{quan2016fusionnet}.
They were trained by a mean square error (MSE) objective between original images and adversarial examples. During training, each input batch consisted of all kinds of adversaries with ratio of FGSM v1 : FGSM v2 : i-FGSM v1 : i-FGSM v2 : mi-FGSM = 0.05 : 0.30 : 0.05 : 0.30 : 0.30. 
FGSM and i-FGSM v1 are slightly perturbed dataset while FGSM and i-FGSM v2 are more heavily perturbed dataset.
To compute MSE, we set the denoisers to generate the same output size (64$\times$64) to the adversarial input. The FusionNet was trained for 300 epochs with Adam optimizer with an initial learning rate of 1.0e$^{-5}$ and L2 regularization coefficient of 1.0e$^{-5}$. The U-Net was trained using fine-tuning with Adam optimizer with an initial learning rate of 1.0e$^{-7}$ and L2 regularization coefficient of 1.0e$^{-6}$. Primary and reduced errors between the training set and corresponding adversaries measured by MSE for each denoiser are shown in Table 4.

\begin{table}[h!]
\centering
\begin{tabular}{l|l|l|l} 
\toprule
          & \begin{tabular}[c]{@{}l@{}}Primary\\\footnotesize{MSE($x_{ori}^{tr}$,}\\   \footnotesize{$~~~~~~~~~~~x_{adv}^{tr}$)}\\ \end{tabular} & \begin{tabular}[c]{@{}l@{}}FusionNet\\\footnotesize{MSE($x_{ori}^{tr}$,}\\   \footnotesize{$~~~~~$DN$_{f}$($x_{adv}^{tr}$))}~~\end{tabular} & \begin{tabular}[c]{@{}l@{}} U-Net\\\footnotesize{MSE($x_{ori}^{tr}$,}\\   \footnotesize{$~~$DN$_{u}$($x_{adv}^{tr}$))}~~\end{tabular}  \\ 
\hline
FGSM v1   & 0.051                                                      & 0.230                                                           &    0.168                                                           \\
FGSM v2   & 13.439                                                     & 7.395                                                            &      8.537                                                         \\
i-FGSM v1 & 1.224                                                      & 0.829                                                            &       0.738                                                        \\
i-FGSM v2 & 4.220                                                      & 1.632                                                            &     1.538                                                          \\
mi-FGSM   & 1.466                                                      & 0.612                                                            &       0.572                                                        \\
\bottomrule
\end{tabular}
\caption{Primary and reduced errors between the training set and corresponding adversaries measured by MSE. $x_{ori}^{tr}$ and $x_{adv}^{tr}$ mean original images and adversarial examples for the training set. DN$_{f}(\cdot)$ and DN$_{u}(\cdot)$ are outputs from trained FusionNet and U-Net denoisers, respectively. Unit is pixels.}
\end{table}

\section{Experiments}
We assume that the prepared classifiers are nearly generalized to the training set with their training accuracy. So, observing the propagation of perturbations for its corresponding adversarial sets is justified, since they can purely contribute to making the classifiers fool. In the remaining part of the paper, please note that we only use a training set and its corresponding adversarial sets.

As seen in Figure \ref{fig:one}, to observe the propagation of adversarial perturbation, we should measure the difference between the feature maps extracted from the identical classifier, layer by layer. We can consider two feature maps; one is extracted by the original input (Baseline), and the other by the adversarial input (Configuration 1). The difference is measured by Euclidean and cosine distance (COS-D) between two feature vectors at the same layer, respectively. Euclidean value is normalized by the number of elements of the feature map, so we call it normalized Euclidean distance (NE-D). Additionally, as the classifiers show limited improvement even though passing through the denoising process (Table 5), we also need to observe the feature maps by denoised adversaries passed by denoiser (Configuration 2). 

\begin{table*}[h!]
\footnotesize
\centering
\begin{tabular}{l|lll|lll|lll|lll|lll} 
\toprule
\multirow{2}{*}{} & \multicolumn{3}{l|}{ResNet V2-50}                                & \multicolumn{3}{l|}{VGG-19}                                      & \multicolumn{3}{l|}{SENet-154~}                                  & \multicolumn{3}{l|}{\begin{tabular}[c]{@{}l@{}}Inception-\\ResNet V2\end{tabular}} & \multicolumn{3}{l}{DenseNet-201}                                  \\ 
\cline{2-16}
                  & \begin{tabular}[c]{@{}l@{}}w/o\\DN \end{tabular} & w/ FN & w/ UN & \begin{tabular}[c]{@{}l@{}}w/o\\DN \end{tabular} & w/ FN & w/ UN & \begin{tabular}[c]{@{}l@{}}w/o\\DN \end{tabular} & w/ FN & w/ UN & \begin{tabular}[c]{@{}l@{}}w/o\\DN \end{tabular} & w/ FN & w/ UN                   & \begin{tabular}[c]{@{}l@{}}w/o\\DN \end{tabular} & w/ FN & w/ UN  \\ 
\hline
FGSM v1           & 98.9                                             & 98.6  & 99.3  & 95.7                                             & 96.3  & 96.0  & 97.3                                             & 96.3  & 96.8  & 91.2                                             & 97.3~ & 98.4                    & 98.1                                             & 98.4  & 97.1   \\
FGSM v2           & 33.8                                             & 64.5  & 69.2  & 27.7                                             & 48.9  & 52.0  & 35.1                                             & 52.5  & 56.6  & 34.2                                             & 57.0  & 61.3                    & 32.0                                             & 60.7  & 60.2   \\
i-FGSM v1         & 77.7                                             & 87.3  & 88.2  & 52.2                                             & 64.7  & 67.9  & 68.7                                             & 74.6  & 76.8  & 71.5                                             & 78.4  & 79.8                    & 69.2                                             & 82.3  & 82.5   \\
i-FGSM v2         & 41.1                                             & 69.2  & 71.1  & 32.2                                             & 48.2  & 48.8  & 39.4                                             & 54.7  & 55.6  & 39.1                                             & 59.2  & 60.8                    & 38.9                                             & 62.2  & 63.7   \\
mi-FGSM           & 94.1                                             & 96.9  & 97.0  & 93.1                                             & 96.3  & 96.1  & 94.6                                             & 95.8  & 96.3  & 94.6                                             & 96.8  & 96.8                    & 94.3                                             & 97.4  & 96.8   \\
\bottomrule
\end{tabular}
\caption{Classification accuracy on the adversarial sets of each classifier with/without the denoiser. The classifiers are generalized to the training set. All classifiers show somewhat reasonable performance to the adversarial sets with small perturbation (FGSM v1, i-FGSM v1, and mi-FGSM). However, they show relatively lower performance to the bigger adversarial sets (FGSM v2, i-FGSM v2) even though inputs pass through the denoising process. FN: FusionNet, UN: U-Net}
\end{table*}

\begin{figure*}[h!]
\begin{center}
   \includegraphics[width=0.8\linewidth]{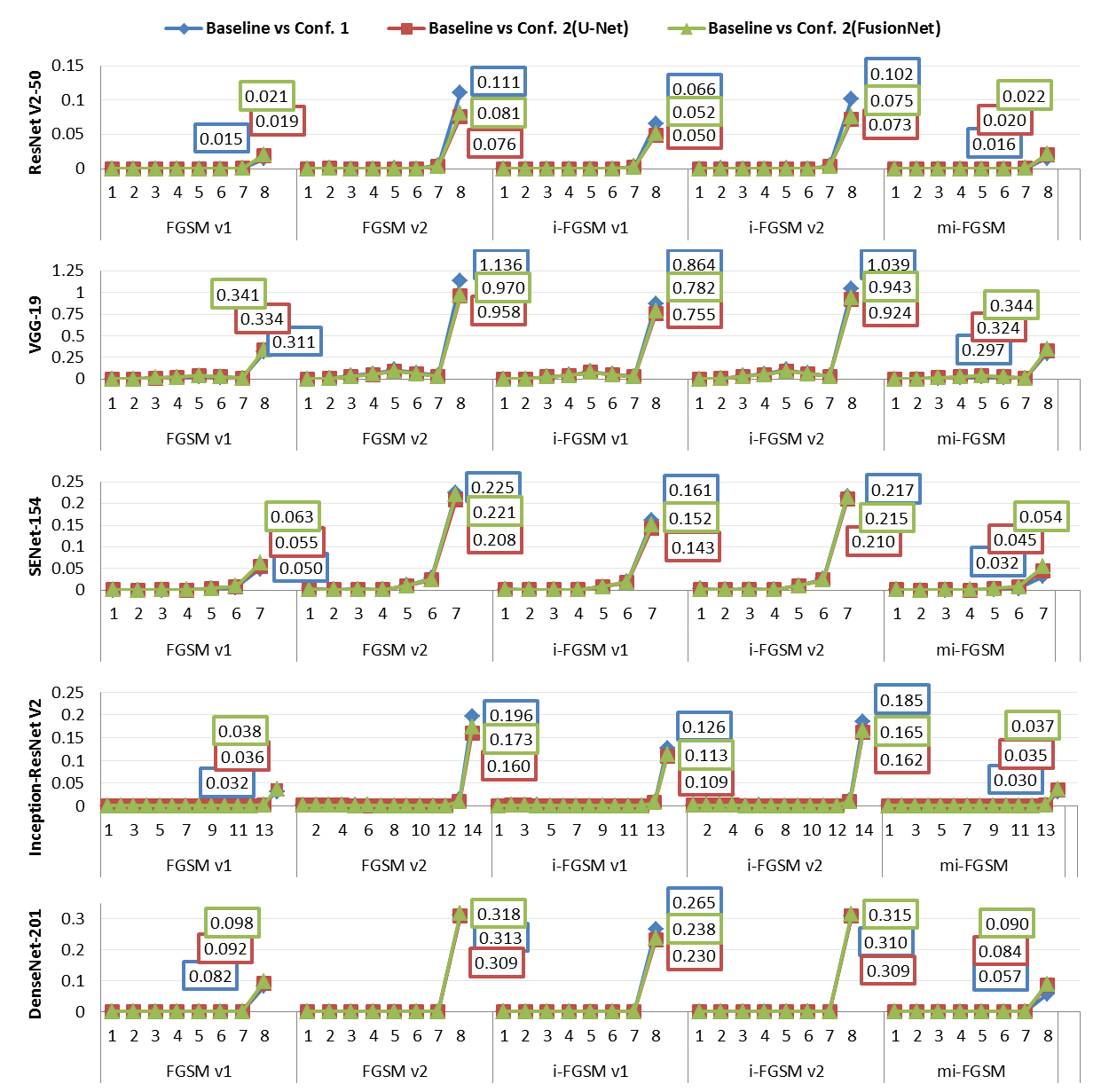}
\end{center}
   \caption{Average normalized Euclidean distance (NE-D) between the feature maps by baseline and configuration settings according to the adversaries and the classifiers. Each figure means the distance at the last observed feature map (i.e. the last layer).}
\label{fig:two}
\end{figure*}

\begin{figure*}[h!]
\begin{center}
   \includegraphics[width=0.8\linewidth]{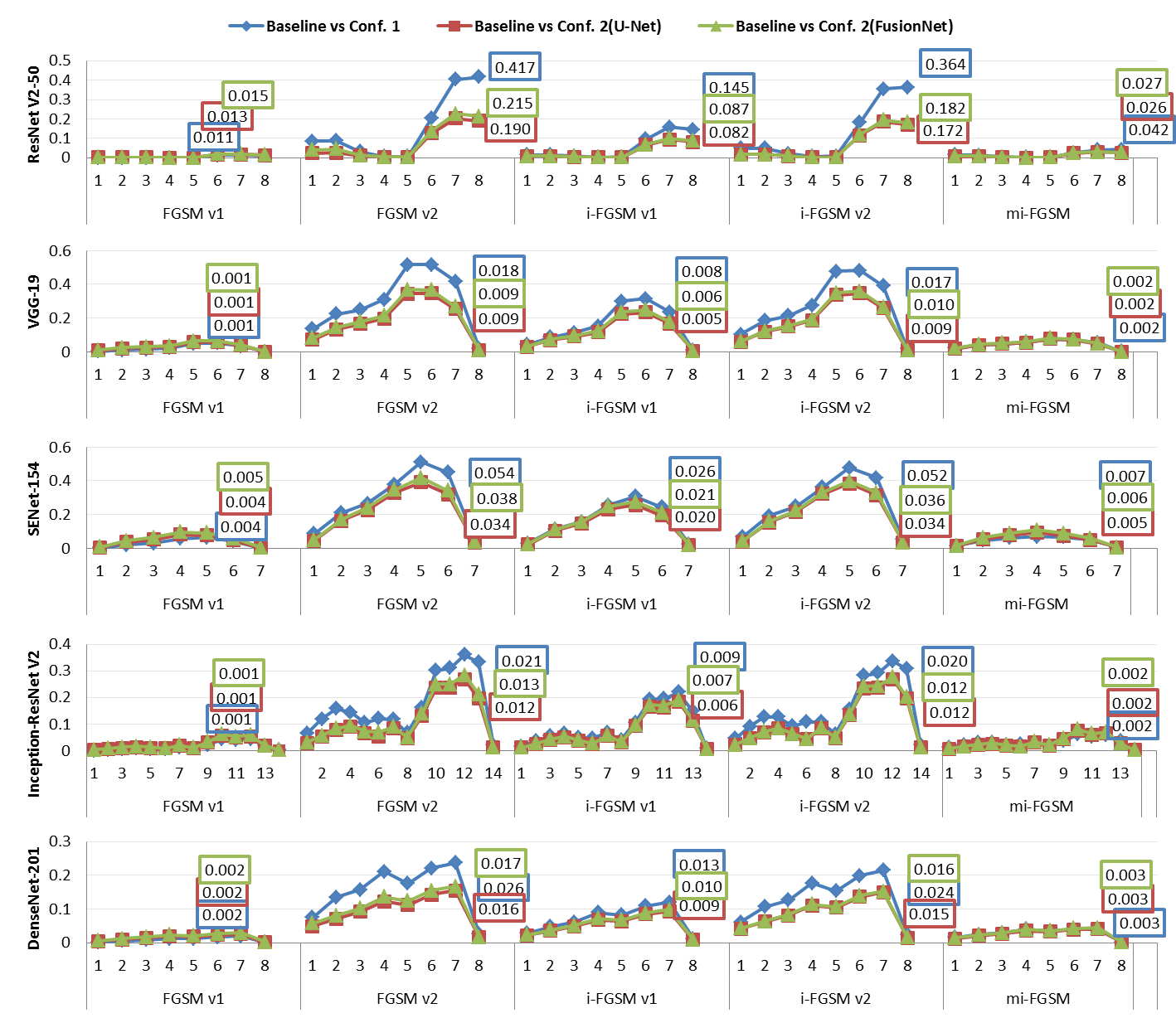}
\end{center}
   \caption{Average COS-D between the feature maps output by baseline and configuration settings according to the adversaries and the classifiers. Each figure is the distance at the last observed feature map (i.e. the last layer).}
\label{fig:three}
\end{figure*}

\begin{figure}[h!]
\begin{center}
\subfloat(a)Radial discriminant hyperplane{
	\label{subfig:foura}
	\includegraphics[width=0.86\linewidth]{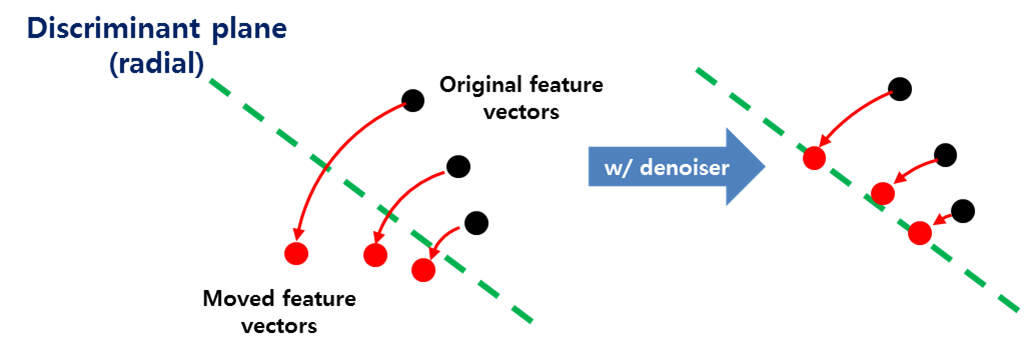}} 
 
\subfloat(b)Tangential discriminant hyperplane{
	\label{subfig:fourb}
	\includegraphics[width=0.86\linewidth]{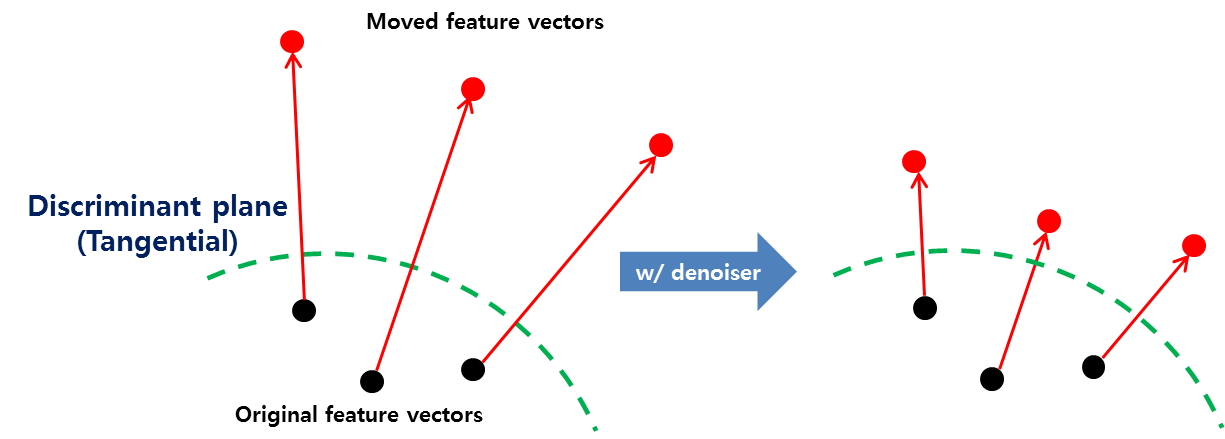}}
\end{center} 
\caption{Two types of discriminant hyperplane. Based on the observations, ResNet V2-50's plane is anticipated as radial and the others' as tangental.}
\label{fig:four}
\end{figure}

\begin{table}[h]
\centering
\scriptsize
\begin{tabular}{l|l|l|l|l|l} 
\toprule
   & Resnet V2-50  & VGG-19  & SENet-154   & \begin{tabular}[c]{@{}l@{}}Inception\\-ResNet V2 \end{tabular}   & \begin{tabular}[c]{@{}l@{}}DenseNet\\~ ~ ~ ~ ~-201 \end{tabular}       \\ 
\hline
1  & 1 convs~      & 2 convs & 3 convs     & 1 conv                                                           & \begin{tabular}[c]{@{}l@{}}1 Conv,\\ 1 max pool,\\ 6 FCs\end{tabular}  \\ 
\hline
2  & 1 max pool    & 2 convs & 3 SE block  & 1 conv                                                           & 1 transition                                                           \\ 
\hline
3  & 3 bottlenecks & 4 convs & 8 SE block  & 1 conv                                                           & 12 FCs                                                                 \\ 
\hline
4  & 4 bottlenecks & 4 convs & 36 SE block & 1 conv                                                           & 1 transition                                                           \\ 
\hline
5  & 6 bottlenecks & 4 convs & 3 SE block  & 1 conv                                                           & 48 FCs                                                                 \\ 
\hline
6  & 3 bottlenecks & 1 FC    & 1 avg pool  & \begin{tabular}[c]{@{}l@{}}1 maxpool,\\ 1 mixed\_b \end{tabular} & 1 transition                                                           \\ 
\hline
7  & 1 avg pool~~  & 1 FC    & 1 FC        & 10 Block35s                                                      & 32 FCs                                                                 \\ 
\hline
8  & 1 FC layer    & 1 FC    &             & mixed\_a                                                         & \begin{tabular}[c]{@{}l@{}}1 max pool,\\1 FC\end{tabular}              \\ 
\hline
9  &               &         &             & 20 Block17s                                                      &                                                                        \\ 
\hline
10 &               &         &             & mixed\_a                                                         &                                                                        \\ 
\hline
11 &               &         &             & 9 Block8s                                                        &                                                                        \\ 
\hline
12 &               &         &             & 1 Block8                                                         &                                                                        \\ 
\hline
13 &               &         &             & \begin{tabular}[c]{@{}l@{}}1 conv,\\ 1 avg pool~\end{tabular}    &                                                                        \\ 
\hline
14 &               &         &             & 1 FC                                                             &                                                                        \\
\bottomrule
\end{tabular}
\caption{Selected feature map lists of each classifier. Feature maps are extracted right after each procedure, which is written in the table.}
\end{table}

Practically, it is time-consuming to observe all of feature maps to all the datasets, due to tremendously large size of them and computational cost. Thus, we appointed some representative feature maps to observe for each CNN and Table 6 shows them. 
Because recently proposed DNNs have too much feature
maps to observe all of them. So we only evaluated the last layers
of each block as a representative layer. It is a common
approach to analyze feature maps, such as \cite{kimin-lee}.
Additionally, as extracting feature maps for all of the image set was burdensome work in the aspect of computational time and resources, we randomly sampled 1,000 images from the original training set and took adversarial examples corresponding to them. Consequently, at one observed position in one CNN, we measured the average distance of 1,000 feature map pairs of the originals (Baseline) and adversaries (Configuration 1) or denoised adversaries (Configuration 2).

In this experiment, PyTorch 0.4.1 and Python 3.5.2 were used in the Ubuntu 16.04 LTS. 

\section{Results and Discussion}

Figure \ref{fig:two} and \ref{fig:three} show experimental results about averaged NE-D and COS-D between the feature maps of baseline and configuration settings for each classifier and adversarial set. In all classifier with/without denoiser, NE-D explosively increases in the fully connected (FC) layer while it tends to keep in small in the convolutional layers (i.e. all layers except for the last one). Generally, adversarial set which has relatively larger perturbation (FGSM v2 and i-FGSM v2) shows a bigger gap than others. Especially, VGG-19 shows an incredibly large jump in the gap after the last FC layer. On the other hand, Figure \ref{fig:three} shows dramatic directional identification at the end of each network by the FC layers, while the directional gap is gradually increased (ResNet V2-50, VGG-19, and SENet-154) or shows up-down-up (Inception-ResNet V2 and DenseNet-201) in the convolutional layers. 
Figure \ref{fig:three}'s result is consistent with \cite{lee2018simple}, which confirmed that a feature vector of the middle layer shows the behavior of outlier.

In the aspect of the effect of denoisers, MSE was not perfectly eliminated by the denoisers (see Table 4), so that they gave not enough improvement in validation accuracy in spite of reduced adversarial noise (see Table 5). As seen the graph in Figure \ref{fig:two} and \ref{fig:three}, NE-D is a little, but not enough, decreased compared to the case without the denoisers (Of course, in ResNet-V2 50, noise is reduced 20 $\sim$ 30 $\%$ in the case of FGSM v2 and i-FGSM v2). COS-D also tends to be reduced during the convolutional layers, but it becomes identical eventually. Based on that observation, denoisers are somewhat effective in reducing COS-D in the middle layers. We should note that NE-D is much a little reduced at the last layers in SENet-154, Inception-ResNet V2 and DenseNet-201. This phenomenon seems that it relates to weaker improvement than the case of ResNet V2-50.

Overall, observations for the experiment follow below,
\begin{enumerate}
    \item In convolutional layers, adversarial perturbation is not amplified with averaged NE-D, but with COS-D. However, in FC layers, it shows the opposite pattern. As an exceptional case, in ResNet V2-50, COS-D is not decreased when passing through the FC layer while it shows relatively lower NE-D than others. 
    \item As seen in Table 5, accuracy is most improved when a denoiser is combined with ResNet V2-50. Seemingly, ResNet V2-50 has successful noise suppression capacity than other classifiers when seen NE-D. However, it poorly controls the direction of noise in the aspect of COS-D. 
    \item Effect of denoiser is different depends on classifiers, but it is limited from the aspect of a logit vector. NE-D shows that the amount of change of ResNet V2-50 is larger than other classifiers, such as SENet-154, Inception-ResNet V2 and DenseNet-201. Denoiser also reduce COS-D in middle layers for all classifiers.
\end{enumerate}

When seen the results of the last FC layer, adversaries result in angular (ResNet V2-50) or longitudinal perturbations (the others). This means that the different discriminant hyperplane might be constructed according to the classifiers. Since the adversarial examples are generated to make the classifiers fool, it is reasonable for us to infer where the discriminant hyperplane is, based on moved feature vectors. When inferring from relative small NE-D and large COS-D, a radial discriminant plane might be reasonable in the case of ResNet V2-50, as seen in Figure \ref{fig:four}(a). On the other hand, in the case of the other classifiers, a tangential one is strongly suspected, as seen in Figure \ref{fig:four}(b). With this inference, it is convincible that effectiveness of the denoisers is maximized in ResNet V2-50, since reduced COS-D (approx. 50$\%$ or more) and small NE-D are very effective to prevent misclassification. On the contrary, in the other cases which have a tangential plane, in spite of tiny COS-D, an improvement on performance cannot easily be achieved unless the denoiser reduces NE-D much a lot. 
Thus, ResNet V2-50 architecture is more efficient than other architectures to reduce NE-D in the experiment.
However, because the adversarial perturbed image set is generated by ResNet only, additional experiments using various attacker have to be conducted for consistency. 

\begin{figure*}[h!]
\begin{center}
   \includegraphics[width=0.8\linewidth]{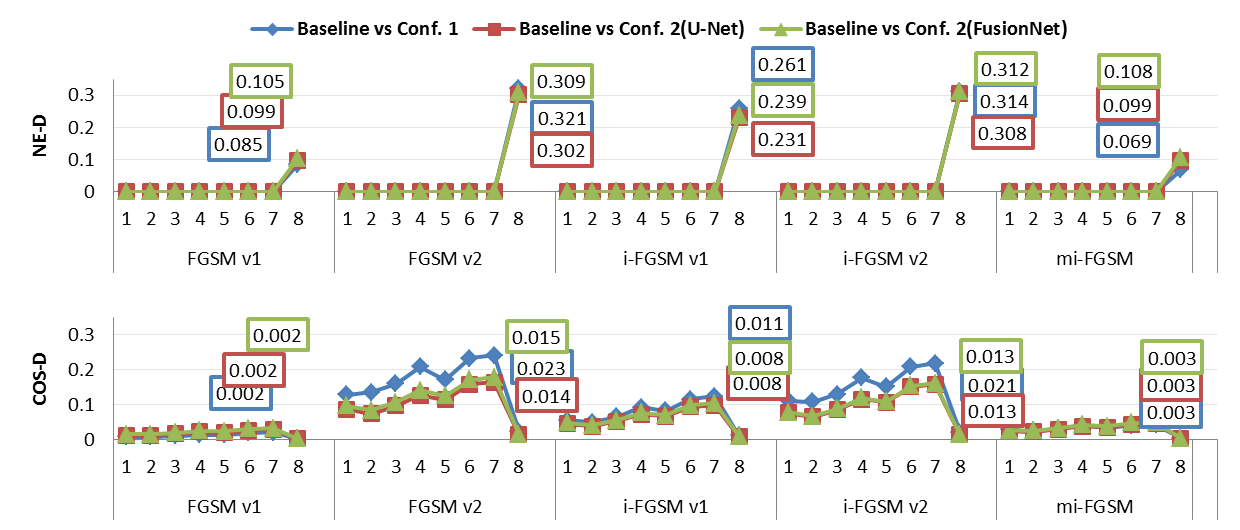}
\end{center}
   \caption{Average NE-D and COS-D in the case of DenseNet-201 which replaced FC to Conv layer.}
\label{fig:five}
\end{figure*}

Due to surprisingly amplified NE-D in the last FC layer, it is natural to consider replacement of it to convolutional one. 
Because DNN is generally trained to be overconfident \cite{guo2017calibration}, it is guessable that final FC layer is trained to make it's distance large.
So, we additionally conducted the same experiment with modified DenseNet-201(i.e. the last layer is modified from avg. pooling(kernel size=4, stride=1, no padding) + FC to Conv. layer (kernel size=4, stride=1, no padding), so the network has only Conv. layers). But, there is no big difference (but, tiny improvement) when comparing to the previous result (See Figure \ref{fig:five}). It is somewhat guessable because it just changes a linear operation from on 1$\times$1 map to on 4$\times$4 that is similar setting in that it refers to whole map, not local. Consequently, the FC layer itself, at least, is not a direct factor of amplification of NE-D.

In fact, except for the last FC layer, all feature maps were observed after Relu activation which reduces NE-D via rectifying the output. At that point, we wondered whether getting lower NE-D is possible or not if Relu is added after the FC. With that setting, we experimented again with DenseNet-201. However, generalization could not be achieved with the training set. So, we changed Relu to PRelu($\alpha = 0.25$) \cite{DBLP:journals/corr/HeZR015} instead. 

\begin{figure*}[h!]
\begin{center}
   \includegraphics[width=0.8\linewidth]{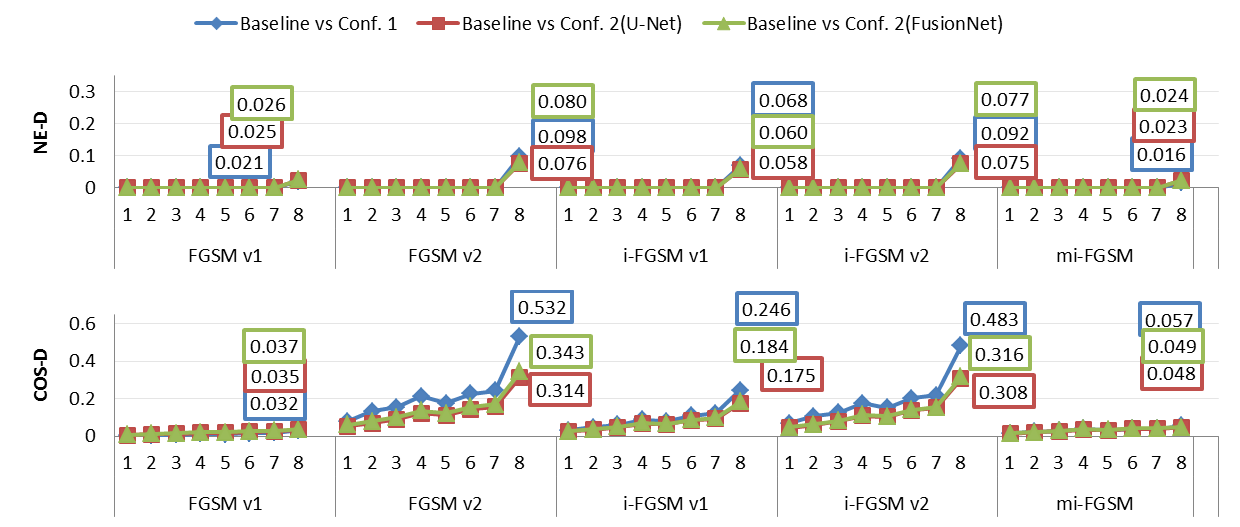}
\end{center}
   \caption{Average NE-D and COS-D in the case of DenseNet-201 adding PRelu after the FC layer.}
\label{fig:six}
\end{figure*}

As a result, NE-D of the model is decreased while COS-D is increased on the opposite side (Figure \ref{fig:six}).
The plot of NE-D and COS-D for the model follows a similar pattern to that of ResNet V2-50, but it little affects the performance (Table 7). 
Consequently, a classifier which is trained to reduce NE-D causes large COS-D logit, and training only using NE-D could not affect the performance. 
Thus, the performance of denoiser for the model is less than that of ResNet V2-50. 

\begin{table}[t!]
\centering
\scriptsize
\begin{tabular}{l|lll|lll} 
\toprule
\multirow{3}{*}{} & \multicolumn{6}{l}{DenseNet-201}                                              \\ 
\cline{2-7}
                  & \multicolumn{3}{l|}{FC-\textgreater{}Conv.} & \multicolumn{3}{l}{adding PRelu after FC}  \\ 
\cline{2-7}
                  & w/o DN & w/ FN & w/ UN                      & w/o DN & w/ FN & w/ UN          \\ 
\hline
FGSM V1           & 97.4   & 96.3  & 96.9                       & 97.8   & 97.6  & 97.8           \\
FGSM V2           & 31.6   & 56.7  & 56.6                       & 30.7   & 59.9  & 59.6           \\
i-FGSM V1         & 63.7   & 74.6  & 73.8                       & 66.3   & 78.2  & 78.7           \\
i-FGSM V2         & 37.1   & 58.2  & 56.4                       & 38.3   & 61.6  & 59.9           \\
mi-FGSM           & 93.9   & 96.0  & 96.8                       & 94     & 95.5  & 96.7           \\
\bottomrule
\end{tabular}
\caption{Accuracy of two types of  modified DenseNet-201s.}
\end{table}


\section{Conclusion}
To defense adversarial attack, the use of denoiser is the most widely used solution. It reduces the amount of noise, but the improvement of classification accuracy is marginal. 
In this paper, we aimed to examine the propagation of adversarial perturbation by measuring Euclidean distance and cosine distance in each CNN layer between each feature map of the original image and perturbed image passed through denoiser.
We observed that Euclidean distance explosively increases in final FC layer while cosine distance fluctuated and disappeared at the last layer in most cases except the ResNet V2-50 classifier. 
In the case of ResNet V2-50, COS-D explosively increased in final FC layer while NE-D disappeared at the last layer. 
The accuracy improvement of ResNet V2-50 is more than that of other networks. 
This means that the two types of distance could be utilized to examine how noise is propagated through the network. 
It would be interesting future work to analysis why the ResNet V2-50 is robust for an adversarial attack.

{\small
\bibliographystyle{ieee}
\bibliography{egbib}
}

\end{document}